\documentclass[times,twocolumn,final]{elsarticle}
\usepackage[table]{xcolor}
\usepackage{cag}
\usepackage{amssymb}
\usepackage{graphicx}
\usepackage{arydshln} 
\usepackage{multirow}
\usepackage{booktabs}
\usepackage{subcaption}
\usepackage{hyperref}
\hypersetup{
    colorlinks=true,
    linkcolor=blue,
    filecolor=magenta,      
    urlcolor=cyan,
}

\usepackage{amsmath}
\usepackage{amsfonts}
\usepackage{orcidlink}

\usepackage{colortbl}
\definecolor{lightgray}{gray}{0.9}

\newcommand{\highlight}[1]{\textcolor{black}{#1}}
\newcommand{\highlightx}[1]{\textcolor{black}{#1}}

\def\onedot{.}
\def\eg{\emph{e.g}\onedot} 
\def\ie{\emph{i.e}\onedot} 
\def\cf{\emph{cf}\onedot}

\def\etal{\emph{et al}\onedot}

\journal{Computers \& Graphics}

\begin{document}

\begin{frontmatter}

\title{TextANIMAR: Text-Based 3D Animal Fine-Grained Retrieval}

\author[1,3]{Trung-Nghia Le\orcidlink{0000-0002-7363-2610}} 
\author[4]{Tam V. Nguyen\orcidlink{0000-0003-0236-7992}}
\author[1,3]{Minh-Quan Le\orcidlink{0000-0002-0709-3225}}
\author[1,3]{Trong-Thuan Nguyen\orcidlink{0000-0001-7729-2927}}
\author[1,3]{Viet-Tham Huynh\orcidlink{0000-0002-8537-1331}}
\author[1,3]{Trong-Le Do\orcidlink{0000-0002-2906-0360}}
\author[1,3]{Khanh-Duy Le\orcidlink{0000-0002-8297-5666}}
\author[1,3]{Mai-Khiem Tran\orcidlink{0000-0001-5460-0229}}
\author[1,3]{Nhat Hoang-Xuan\orcidlink{0000-0001-8759-0354}}
\author[1,3]{Thang-Long Nguyen-Ho\orcidlink{0000-0003-1953-7679}}
\author[2,3]{Vinh-Tiep Nguyen\orcidlink{0000-0003-4260-7874}}
\author[1,3]{Tuong-Nghiem Diep}
\author[1,3]{Khanh-Duy Ho}
\author[1,3]{Xuan-Hieu Nguyen}
\author[1,3]{Thien-Phuc Tran}
\author[1,3]{Tuan-Anh Yang}
\author[1,3]{Kim-Phat Tran}
\author[1,3]{Nhu-Vinh Hoang}
\author[1,3]{Minh-Quang Nguyen}
\author[1,3]{E-Ro Nguyen}
\author[1,3]{Minh-Khoi Nguyen-Nhat}
\author[1,3]{Tuan-An To}
\author[1,3]{Trung-Truc Huynh-Le}
\author[1,3]{Nham-Tan Nguyen}
\author[1,3]{Hoang-Chau Luong}
\author[1,3]{Truong Hoai Phong}
\author[1,3]{Nhat-Quynh Le-Pham}
\author[1,3]{Huu-Phuc Pham}
\author[1,3]{Trong-Vu Hoang}
\author[1,3]{Quang-Binh Nguyen}
\author[1,3]{Hai-Dang Nguyen\orcidlink{0000-0003-0888-8908}}
\author[5]{Akihiro Sugimoto\orcidlink{0000-0001-9148-9822}}
\author[1,3]{Minh-Triet Tran\orcidlink{0000-0003-3046-3041}\corref{cor1}} \ead{tmtriet@fit.hcmus.edu.vn}

\affiliation[1]{organization={University of Science, VNU-HCM}, city={Ho Chi Minh City}, country={Vietnam}}
\affiliation[3]{organization={Vietnam National University}, city={Ho Chi Minh City}, country={Vietnam}}
\affiliation[4]{organization={University of Dayton}, city={Ohio}, country={U.S.}}
\affiliation[2]{organization={University of Information Technology, VNU-HCM}, city={Ho Chi Minh City}, country={Vietnam}}
\affiliation[5]{organization={National Institute of Informatics}, city={Tokyo}, country={Japan}}

\cortext[cor1]{Corresponding author}

\received{\today}




\begin{abstract}

3D object retrieval is an important yet challenging task that has drawn more and more attention in recent years. While existing approaches have made strides in addressing this issue, they are often limited to restricted settings such as image and sketch queries, which are often unfriendly interactions for common users. In order to overcome these limitations, this paper presents a novel SHREC challenge track focusing on text-based fine-grained retrieval of 3D animal models. Unlike previous SHREC challenge tracks, the proposed task is considerably more challenging, requiring participants to develop innovative approaches to tackle the problem of text-based retrieval. Despite the increased difficulty, we believe this task can potentially drive useful applications in practice and facilitate more intuitive interactions with 3D objects. Five groups participated in our competition, submitting a total of 114 runs. While the results obtained in our competition are satisfactory, we note that the challenges presented by this task are far from fully solved. As such, we provide insights into potential areas for future research and improvements. We believe we can help push the boundaries of 3D object retrieval and facilitate more user-friendly interactions via vision-language technologies.

\end{abstract}


\end{frontmatter}


\section{Introduction}
\label{sec:introduction}

The rapid growth and advancement of 3D technologies have significantly expanded the availability and abundance of 3D objects. As a result, 3D object retrieval has emerged as a prominent and interesting research area, with substantial practical applications~\cite{stotko2019slamcast,liu2019real, wang2020rgb2hands, guo2023vid2avatar, koca2019augmented} across diverse domains, including video games, creative arts, motion picture production, and virtual reality.

In real-world scenarios, obtaining a 3D model as a query typically demands significant effort and resources. As a solution, content-based 3D object retrieval techniques~\cite{he2018triplet, li2020mpan, kim2021method} have been developed, which provide a more accessible approach to query collection. These techniques aim to retrieve 3D objects from a database based on their visual content, encompassing color, texture, shape, and geometric features. Among the various retrieval methods, image-based and sketch-based approaches~\cite{han2019image, song2022gradual, song2023self, nie2023image} have gained popularity. Image-based retrieval methods leverage RGB images captured from the real world to extract relevant visual features, facilitating the retrieval of similar 3D models. This approach is advantageous as it allows users to access valuable 3D models conveniently through readily available 2D images. In contrast, sketch-based retrieval~\cite{Qin-SHREC2022, shi2021rebor, yang2022sequential, bai2023hda2l} employs hand-drawn sketches as queries. The intuitive nature of freehand drawings allows for a more effective capture of the essential features of 3D objects while filtering out irrelevant information. Nevertheless, image-based and sketch-based approaches introduce notable challenges in 3D object retrieval research. The substantial disparities between 2D and 3D modalities present a significant obstacle, as 2D images or sketches differ considerably from their corresponding 3D counterparts and perspectives. Moreover, sketches are prone to ambiguity and errors, which can detrimentally affect the accuracy of the retrieval process. 

We have introduced a novel challenge track called \textit{\textbf{Text}-based 3D \textbf{ANIMA}l model fine-grained \textbf{R}etrieval (\textbf{TextANIMAR})}\footnote{\url{https://aichallenge.hcmus.edu.vn/textanimar}} aiming to enhance the effectiveness of content-based 3D object retrieval. The primary objective of this track is to retrieve relevant 3D animal models from a dataset using textual queries. This SHREC challenge track poses significantly greater challenges and provides a more effective simulation of real-life scenarios than previous SHREC challenge tracks. \highlight{We also note that after the challenge concluded, the dataset has been made publicly available for academic purposes.} 

Firstly, in conventional 3D object retrieval tasks, the primary focus is typically on the object category. \highlight{These approaches often involve training and testing with samples from the same category. While this leads to feature extraction methods specialized for known categories, it may limit their effectiveness with unseen categories in practice. Their efficacy should be evaluated only within specific contexts.} Nevertheless, alternative approaches, such as open-set 3D object retrieval, have emerged as promising solutions for effectively addressing the retrieval of unseen categories. These approaches involve training models on known-category 3D objects and incorporating unseen-category data, offering potential avenues to overcome the challenges posed by classification-based methods. Regardless, our fine-grained retrieval task requires participants to accurately search 3D animal models whose shapes correspond to a given query, necessitating consideration of unseen categories and poses (\cf~Table~\ref{tab:3d_object_retrieval_tasks}). This task poses a more significant challenge than traditional category-based retrieval, which requires handling the substantial discrepancies in animal breeds and poses.

\highlight{Second, the quality and resolution of the input image can impact the performance of image-based 3D object retrieval. However, controlling these factors requires additional effort. It is also worth noting that image queries may encounter challenges in effectively handling variations in object scale, orientation, and perspective, potentially impacting retrieval performance. Conversely, sketches trained on existing datasets often exhibit semi-photorealistic qualities and are expertly created, posing challenges for regular users to reproduce them in real-world scenarios.} Last but not least, text-based queries are considerably easier to generate than image capture or sketching, making them a more user-friendly alternative. We anticipate that the text-based 3D animal fine-grained retrieval task will stimulate new research directions and find practical applications.

The structure of this paper is as follows. Section~\ref{sec:related_work} discusses the literature review and previous works relevant to 3D object retrieval. Section~\ref{sec:dataset} presents the ANIMAR dataset and the evaluation metrics used in this SHREC challenge track. The participant statistics are reported in Section~\ref{sec:participants}. In Section~\ref{sec:methods}, we describe the methods employed by the participating teams. Section~\ref{sec:results} contains the evaluation results, including a detailed analysis of the performance of the different methods. Finally, in Section~\ref{sec:conclusion}, we summarize the key points of the paper and discuss the implications for future research in this field.

\begin{table}[!t]
\centering
\caption{SHREC challenge tracks for 3D object retrieval.}
\label{tab:3d_object_retrieval_tasks}
\resizebox{\columnwidth}{!}{%
\begin{tabular}{lcccc}
\hline
  \begin{tabular}[c]{@{}c@{}}\textbf{SHREC}\\\textbf{Challenge}\end{tabular} &
  \textbf{Year} &
  \begin{tabular}[c]{@{}c@{}}\textbf{Query}\\\textbf{Type}\end{tabular} &
  \begin{tabular}[c]{@{}c@{}}\textbf{Training}\\\textbf{Category}\end{tabular} &
  \begin{tabular}[c]{@{}c@{}}\textbf{Testing}\\\textbf{Category}\end{tabular}  \\ \hline

Pratikakis~\etal~\cite{Pratikakis-SHREC2016} & 2016 & 3D Shape & Seen & Seen \\ 
Sipiran~\etal~\cite{Sipiran-SHREC2021} & 2021 & 3D Shape & Seen & Seen \\ 

Juefei~\etal~\cite{Juefei-SHREC2018} & 2018 & Sketch & Seen & Seen \\ 
Juefei~\etal~\cite{Juefei-SHREC2019} & 2019 & Sketch & Seen & Seen \\ 
Qin~\etal~\cite{Qin-SHREC2022} & 2022 & Sketch & Seen & Seen \\ 

Hameed~\etal~\cite{Hameed-SHREC2018} & 2018 & Image & Seen & Seen \\ 
Hameed~\etal~\cite{Hameed-SHREC2019} & 2019 & Image & Seen & Seen \\ 
Li~\etal~\cite{Li-SHREC2019} & 2019 & Image & Seen & Seen \\ 
Li~\etal~\cite{Li-SHREC2020} & 2020 & Image & Seen & Seen \\ 
Feng~\etal~\cite{Feng-SHREC2022} & 2022 & Image & Seen & Unseen \\ 

\rowcolor{lightgray} TextANIMAR & 2023 & Text & Unseen & Unseen \\ \hline
\end{tabular}%
}
\end{table}

\section{Related Benchmark}
\label{sec:related_work}

Content-based 3D object retrieval aims to retrieve 3D objects from a database by analyzing the visual contents of the objects, including color, texture, shape, and geometric features. In order to facilitate research in this field, previous SHREC challenges have included various tracks dedicated to related tasks (see Table~\ref{tab:3d_object_retrieval_tasks}).

Few SHREC tracks focus on retrieving 3D objects from a database similar in shape to a given query 3D objects. Pratikakis~\etal~\cite{Pratikakis-SHREC2016} introduced the concept of partial 3D object retrieval, which addresses scenarios where the available information about the query object is incomplete. These techniques help build digital libraries of cultural heritage objects, which require partial 3D object retrieval capabilities. In a related direction, Sipiran~\etal~\cite{Sipiran-SHREC2021} also held a competition to evaluate the ability of retrieval methods to discriminate cultural heritage objects by overall shape.

\begin{figure*}[!t]
     \centering
     \includegraphics[width=\linewidth]{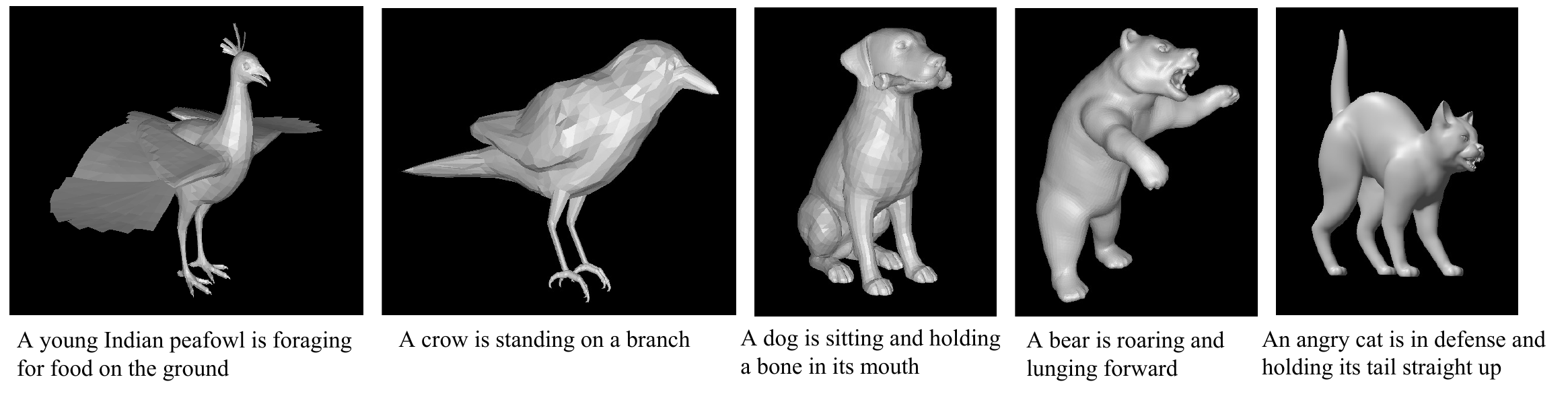}
    \caption{\highlightx{Text ground-truth in ANIMAR dataset, including the text query and the corresponding 3D animal model.}}
    \label{fig:dataset}
    \vspace{-3mm}
\end{figure*}

In contrast, the appeal of sketch-based 3D object retrieval was based on the natural and intuitive nature of freehand sketches and has attracted significant attention in recent years. This area of research has been actively promoted through SHREC tracks organized by Juefei~\etal~\cite{Juefei-SHREC2018, Juefei-SHREC2019} focused on 2D scene sketch-based 3D scene retrieval. To address the domains shift between the sketch and 3D object, domain adaptation (e.g., two-stream CNN with triplet loss, adversarial training, and various data augmentation methods) was employed. In addition, Qin~\etal~\cite{Qin-SHREC2022} further advanced the task by organizing a competition for sketch-based 3D shape retrieval in real-world settings. This competition involved a large-scale collection of sketches drawn by amateurs with varying levels of drawing skills as well as a diverse set of 3D shapes, including models scanned from natural objects. Solutions were developed to simulate realistic retrieval scenarios, incorporating techniques like point cloud and multi-view learning using different deep learning architectures.

Image-based approaches have dominated the field of content-based 3D object retrieval. The SHREC competitions organized by Hameed~\etal~\cite{Hameed-SHREC2018, Hameed-SHREC2019} have significantly advanced 3D scene retrieval from 2D scene image queries. These methods capture different views of 3D scenes for feature learning, incorporating saliency algorithms to select the most promising views for each 3D model. Feature extraction techniques such as Bag of Visual Words have been employed for extracting features from 2D images. Furthermore, VGG, ResNet50, Two-stream CNN, and Conditional Variational autoencoder combined data augmentation demonstrate their effectiveness in this task. Li~\etal~\cite{Li-SHREC2019, Li-SHREC2020} organized SHREC tracks focused on searching for relevant 3D everyday objects using monocular images captured in real-world settings. In these competitions, various deep learning architectures were utilized to learn captured 2D views of 3D objects. 

In conventional 3D object retrieval tasks, all 3D models are categorized, which may not fully capture the diversity present in real-world objects. To address this limitation, recent SHREC tracks proposed by Feng~\etal~\cite{Feng-SHREC2022} have evaluated the performance of different retrieval algorithms under the open-set setting and modality-missing setting. The submitted methods, such as multi-modal learning, have shown promising results in retrieving 3D objects from unknown categories, where the retrieval sets include categories not seen in the training set. However, the open-set retrieval setting still needs to fully simulate the real world when models are trained on known categories. Different from other 3D object retrieval tasks, our TextANIMAR competition stands out by fully simulating real-world scenarios. This is accomplished by utilizing unseen categories for both the train and test 3D objects, providing a more challenging and realistic evaluation setting.

\section{Dataset and Evaluation}
\label{sec:dataset}

 \begin{figure}[!t]
	\centering
\includegraphics[width=\linewidth]{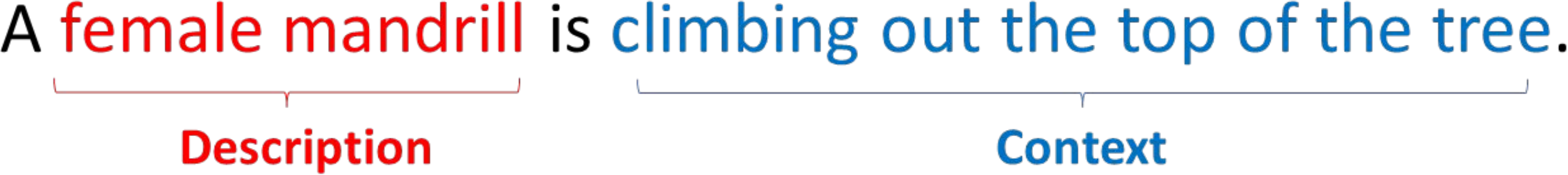}
        \caption{The text query comprises two main components: a description of the animal and a context.}
	\label{fig:text_analysis}
\end{figure}

\subsection{Dataset}

In this competition, we constructed a new dataset, namely ANIMAR, which encompasses a corpus of 711 distinct 3D animal models along with 150 text queries.

We collected 186 mesh models representing over 50 diverse animal categories from Planet Zoo\footnote{\url{https://www.planetzoogame.com}}~\cite{Wu-NeurIPS2022}, a publicly available online resource and video games. Our main objective for this competition track is to imitate real-world scenarios where users seek to explore and identify various types of animals. To this end, we intentionally concealed categorical information throughout the training and retrieval stages. Additionally, we created a simplified set of watertight mesh models by reducing the number of faces by $25\%, 50\%,$, and $75\%$, resulting in a total of 525 models. Following the approach of Douze~\etal~\cite{Douze-2021}, our 3D animal model database is utilized for both the training and retrieval phases.

\highlight{We manually curated 150 English sentences, each incorporating two fundamental constituents: an explicit depiction of the animal's natural shape, focusing on breed-specific attributes, and a context-driven description to match the desired pose for model utilization (see Fig.~\ref{fig:text_analysis}). For instance, when searching for a tiger model suitable for a hunting action, the corresponding description, "a tiger is hunting," ensures the retrieved model possesses the most appropriate pose. Furthermore, we emphasize searching single-animal models where only single-animal descriptions are utilized to optimize the search process. Leveraging these context-aware descriptions enables more accurate and efficient retrieval of 3D animal models, fostering an enhanced user experience in virtual environments. We expect this to facilitate interactive search, whereby users can effortlessly explore and identify 3D animal models based on their species, actions, or even environmental contexts.}

In our dataset, 100 sentences are aligned with their corresponding models in the database resulting in 382 pairs of query-model for training, while the remaining 50 sentences are utilized as queries, corresponding to 188 pairs of query-model, during the retrieval phase. \highlightx{Figure~\ref{fig:dataset} illustrates examples in our ANIMAR dataset, including the text queries and the corresponding 3D animal models.}

\subsection{Evaluation Metrics}

\highlightx{We provide a comprehensive evaluation of the performance of different methods in this track. The following metrics are utilized:}
\begin{itemize}
    \item \textbf{Nearest Neighbor (NN)} evaluates top-1 retrieval accuracy.
    \item \textbf{Precision-at-10 (P@10)} is the ratio of relevant items in the top-10 returned results.
    \item \textbf{Normalized Discounted Cumulative Gain (NDCG)} is a measure of ranking quality defined as $\sum_{i=1}^{p} \frac{rel_i}{log_2(i + 1)}$, where $p$ is the length of the returned rank list, and $rel_i$ denotes the relevance of the $i$-th item.
    \item \textbf{Mean Average Precision (mAP)} is the area under the precision-recall curve that measures the precision of methods at different levels and then takes the average. mAP is calculated as $ \frac{1}{r} \sum_{i=1}^{r} P(i)(R(i) - R(i-1))$, where $r$ is the number of retrieved relevant items, $P(i)$ and $R(i)$ are the precision and recall at the position of the $i^{th}$ relevant item, respectively.
    \item \highlight{\textbf{First Tier (FT)} indicates the recall of the top $m$ retrieval results, where $m$ represents the number of relevant images present in the entire database. It quantifies the accuracy of retrieving the most relevant images among all possible matches. The FT score is calculated by dividing the number of relevant images retrieved in the top $m$ by $m$.}
    \item \highlight{\textbf{Second Tier (ST)} measures the recall of the top $2m$ retrieval results, where $m$ represents the number of relevant images in the entire database. It assesses the system's ability to retrieve relevant images within a broader set of results. The ST score is calculated by dividing the number of relevant images retrieved in the top $2m$ by $m$.}
    \item \highlight{\textbf{Fallout Rate (FR)} reflects the ratio of non-relevant retrieved items to the total number of non-relevant items available. It evaluates the system's effectiveness in avoiding the retrieval of non-relevant items. The FR score is calculated using the formula: dividing the number of non-relevant items retrieved by the total number of non-relevant items.}
\end{itemize}

\section{Participants}
\label{sec:participants}

Five groups participated in the TextANIMAR challenge track, collectively submitting 114 runs. The contest had a three-week duration for participants to complete their submissions. To participate, each group was required to register and submit their results, including a description of the methods employed. It is worth noting that the organizers did not participate in this challenge. Below are the details of the participating groups \highlight{(team members will be added upon acceptance)}:

\begin{itemize}
    \item Polars team submitted by Minh-Khoi Nguyen-Nhat, Tuan-An To, Trung-Truc Huynh-Le, Nham-Tan Nguyen, and Hoang-Chau Luong 
    (see Section \ref{sec:team_polars}).
    \item TikTorch team submitted by Nhat-Quynh Le-Pham, Huu-Phuc Pham, Trong-Vu Hoang, Quang-Binh Nguyen, and Hai-Dang Nguyen 
    (see Section \ref{sec:team_tiktorch}). 
    \item Etinifni team submitted by Tuong-Nghiem Diep, Khanh-Duy Ho, Xuan-Hieu Nguyen, Thien-Phuc Tran, Tuan-Anh Yang, Kim-Phat Tran, Nhu-Vinh Hoang, and Minh-Quang Nguyen 
    (see Section \ref{sec:team_etinifni}). 
    \item THP team submitted by Truong Hoai Phong 
    (see Section \ref{sec:team_thp}).
    \item Nero team submitted by E-Ro Nguyen 
    (see Section \ref{sec:team_nero}).
\end{itemize}

\begin{figure}[!t]
	\centering
	\includegraphics[width=\linewidth]{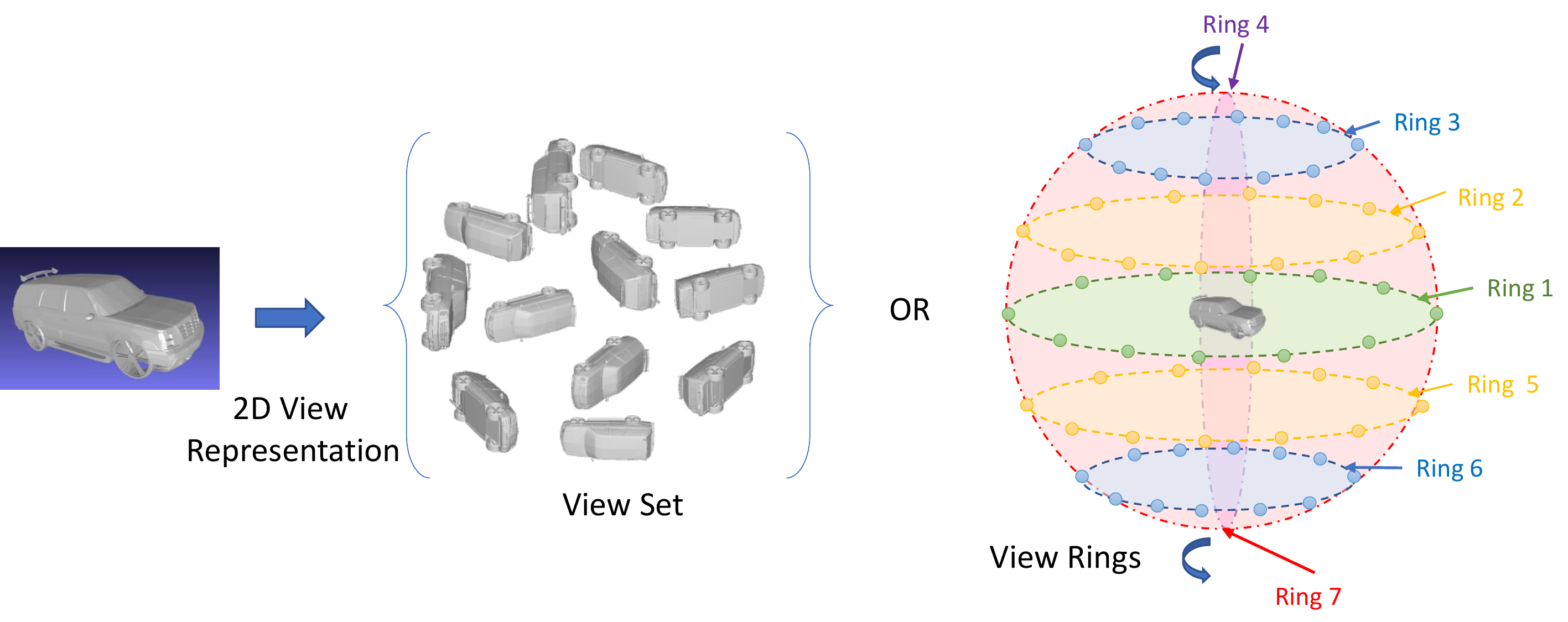}
        \caption{For the 3D object representation, the set of images generated is $R=7$ rings with $V=12$ views on each ring. The chosen latitudes were $0$ (the equator), $\pm 90$ (the poles), and $\pm 30, \pm 60$.}
	\label{fig:object2multiview}
\end{figure}

\section{Methods}
\label{sec:methods}


\subsection{Overview of Submitted Solutions}

\highlight{The solutions submitted to our track can be categorized into two distinct groups; each uses different techniques for representing 3D objects. The former group (\ie, model-based learning approach)  directly learns point cloud, while the latter group (\ie, view-based learning) captures the 3D object as a set of random images.}

\highlight{The model-based learning approach is exemplified by the Polars team. This approach directly learns point clouds via PointNet~\cite{Qi-CVPR2017} and PointMLP~\cite{Ma-2022} to facilitate the representation of 3D animal objects (as shown in Section \ref{sec:team_polars}).}

\begin{figure*}[!t]
\centering
\includegraphics[width=0.82\linewidth]{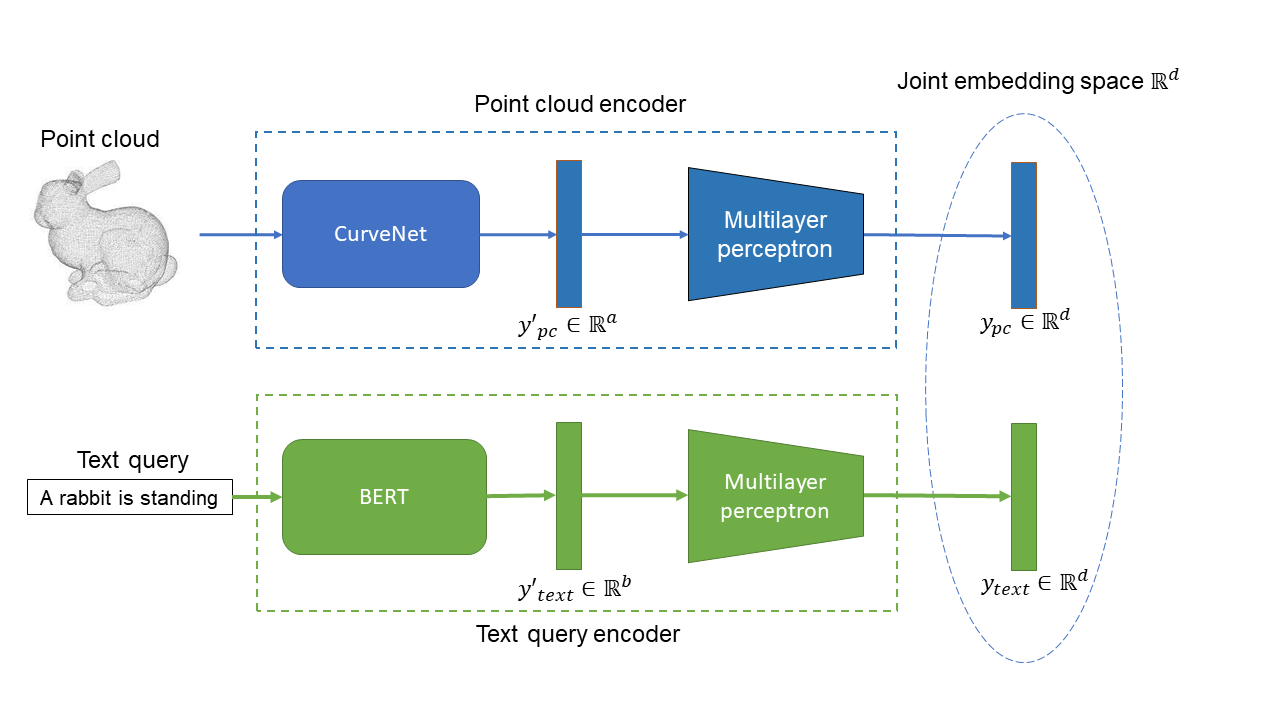}
\caption{Overview of proposed text-cloud contrastive learning framework of Polars team.}
\label{fig:overview_polars}
\end{figure*}

The view-based learning approach involves the collaboration of multiple teams, namely TikTorch, Etinifni, THP, and Nero. This approach represents each 3D object using a series of ring images, as illustrated in Fig.~\ref{fig:object2multiview}. These images are captured by moving a camera around the object along a predefined path, with each ring including a collection of images. The effectiveness of the multi-view technique is particularly notable when the camera follows a trajectory parallel to the ground plane about the object. This approach provides valuable images for learning the distinctive features of 3D objects. \highlight{Although sharing similar concepts, the TikTorch team has developed novel encoders (as depicted in Section \ref{sec:team_tiktorch}). In contrast, the other teams rely on CLIP models~\cite{Radford-ICML2021} (as shown in Sections \ref{sec:team_etinifni},~\ref{sec:team_thp}, and ~\ref{sec:team_nero}) to support their research efforts.}


\begin{figure*}[!t]
	\centering
\includegraphics[width=\linewidth]{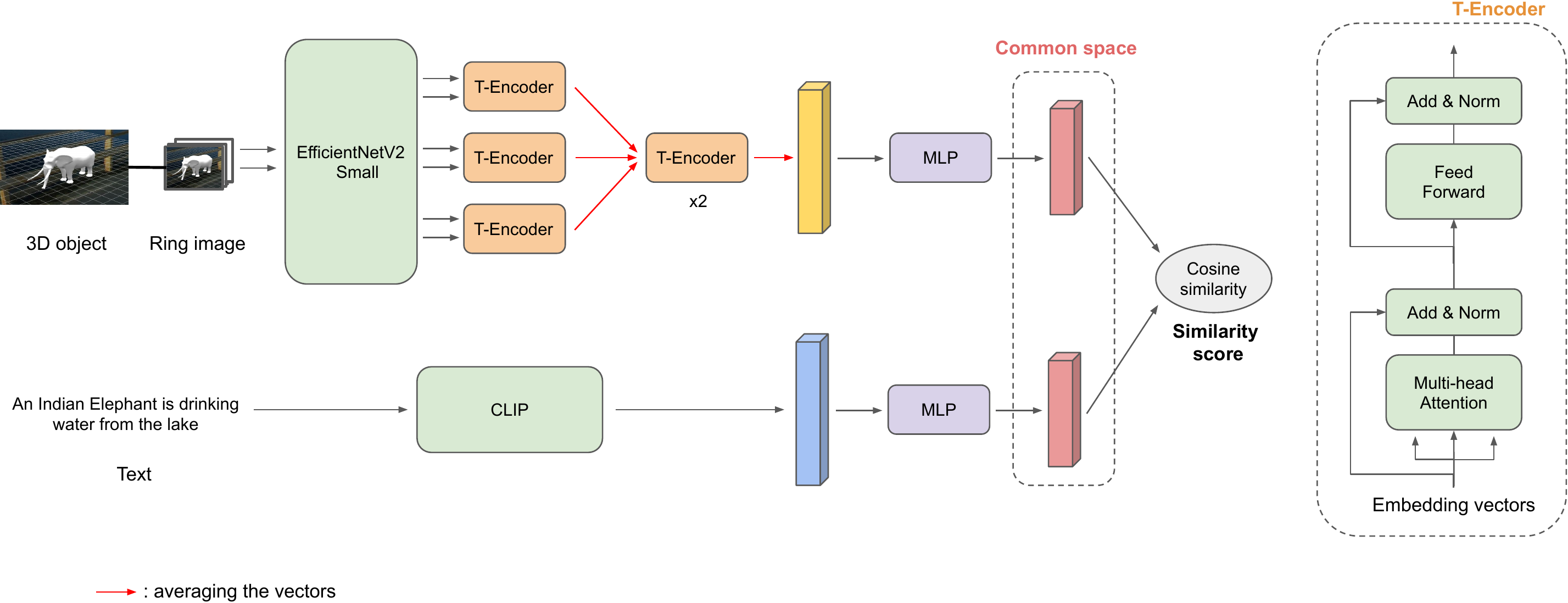}
        \caption{Proposed contrastive learning solution of TikTorch team.}
	\label{fig:overview_tiktorch}
\end{figure*}

\subsection{Polars Team}
\label{sec:team_polars}

\subsubsection{Proposed Framework}

As illustrated in Fig.~\ref{fig:overview_polars}, their proposed framework consists of a query encode branch and a point cloud encode branch. The former encodes the query in natural language into a vector in a joint embedding space, while the latter does the same work for point cloud input. 

The query encoder uses pre-trained BERT~\cite{Devlin-2018} to extract the text query's raw embedding feature. The point cloud encoder employs PointNet~\cite{Qi-CVPR2017} and CurveNet~\cite{Muzahid-JAS2020} to extract the raw embedding feature. Each raw embedding feature is then forwarded into a projection module (\ie, multi-layer perception) in order to map the feature into a joint latent space $\mathcal{R}^{d}$. 

\highlight{In addition, they employ the InfoNCE loss~\cite{Oord-2018} \highlightx{to enhance the learning representation}. Specifically, given a pair of embeddings, ($y_{text}$) for the text query and ($y_{pc}$) for the point cloud, they convert the text query index into an integer ($l$). This allows them to generate two positive pairs for optimization: $(y_{text}, l)$ and $(y_{pc}, l)$. By utilizing these positive pairs, they aim to optimize the representation learning process and improve the overall performance of the system.}

\subsubsection{Training Details}

The based learning rate was $0.001$ and was scheduled by a MultiStepLR at steps $120$, $250$, $350$, and $500$, respectively. A target embedding space dimension $d$ was $128$. In training, they froze the BERT parameters and trained on the remaining part of the model.


\subsection{TikTorch Team}
\label{sec:team_tiktorch}

\subsubsection{Contrastive Learning Solution}

Figure~\ref{fig:overview_tiktorch} illustrates their proposed contrastive learning framework for text-based 3D animal fine-grained retrieval. From two different domains (3D objects and sentences), they try to learn embedding vectors for objects and texts in a common vector space, in which the embedding vectors of similar objects and texts will be closer to each other and vice versa.

To achieve this goal, the team constructs two feature extractors: a 3D Object Feature Extractor and a Text Feature Extractor. These extractors generate two feature vectors, one with $U$ dimensions and the other with $V$ dimensions. Subsequently, these feature vectors are embedded in a shared vector space with $P$ dimensions using two Multi-layer Perceptron (MLP) networks. The contrastive loss~\cite{Chen-ICML2020} is applied to facilitate the simultaneous learning of parameters for both models.

\textbf{3D object feature extractor.} Each 3D object is represented by a collection of seven rings, with each ring consisting of 12 images (\cf~Fig.~\ref{fig:object2multiview}). The team focuses on optimizing the features extracted from the equator ring, as it offers the most significant potential for distinguishing between objects. The extractor module operates a ring extractor extracting the features of the images within each ring and then combining these image features to obtain the overall features of the object.

In the ring extractor, the team performs fine-tuning of EfficientNetV2-Small~\cite{Tan-ICML2021} to extract features from the 12 images within each ring. These extracted feature vectors undergo encoding using the T-Encoder, a part of the Transformer's encoder~\cite{Vaswani-NeurIPS2017}. The T-Encoder captures the relationship among the images within the same ring, determining their relative significance. Subsequently, the feature vectors are combined by taking their average, resulting in a single feature vector for each ring. After obtaining the feature vectors for the three rings, they pass through two T-encoder blocks. Finally, these feature vectors are averaged to derive the overall feature vector representing the 3D object.

\highlight{\textit{Data pre-processing. }}

\begin{figure}[!t]
	\centering
\includegraphics[width=\linewidth]{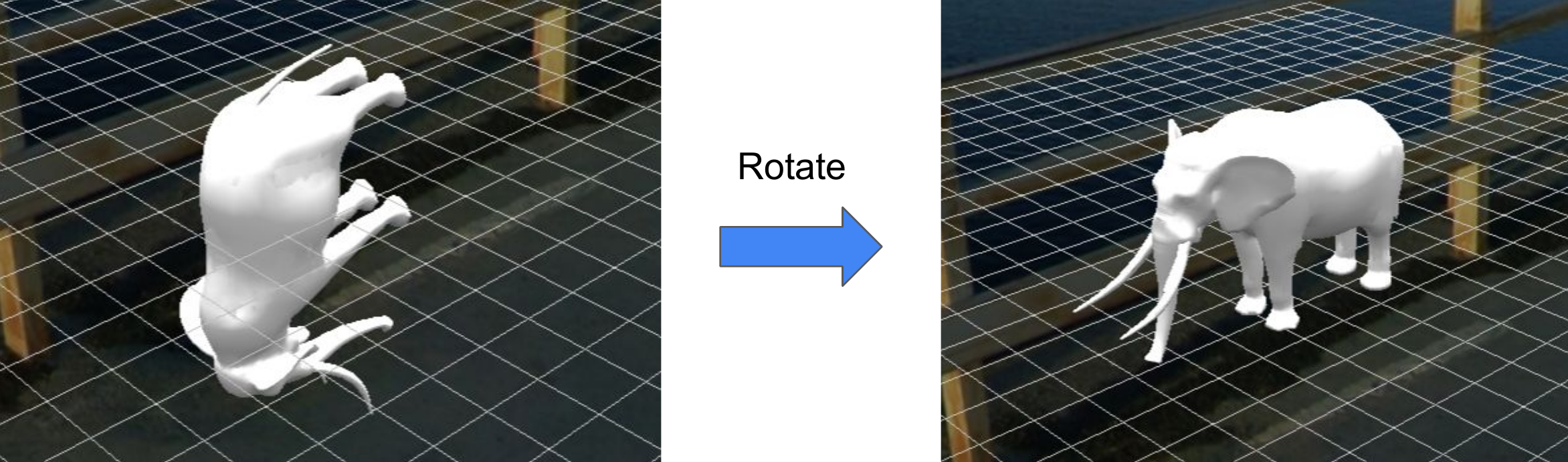}
	\caption{Example of rotating a 3D object whose axis is not aligned with the majority of objects.}
	\label{fig:objectrotation}
\end{figure}

In order to maintain consistency in the resulting multi-view images and their corresponding camera angles, it is crucial to synchronize the axial orientation of the 3D objects before generating batch images. To accomplish this, the team comprehensively examined the available dataset, identifying several objects rotated at a 90-degree angle along the $Ox$ axis. Subsequently, these objects are consistently rotated to align with the majority of the dataset, as illustrated in Fig.~\ref{fig:objectrotation}. They capture images of the objects using a camera angle set to capture images from bottom to top (ring 0 to ring 6), as shown in Fig.~\ref{fig:object2multiview}. They find that the most informative views are captured from ring 3, which provides a direct side perspective from the object. Hence, they focus on processing the images from ring 3 to extract the relevant features and information.

\textbf{Text feature extractor.} To extract the features of the prompt, they fine-tune the CLIP text encoder~\cite{Radford-ICML2021}, a masked self-attention Transformer. This encoder was pre-trained to maximize the similarity of pairs of image and text via a contrastive loss. The models in the CLIP family reduce the parameter size significantly while maintaining competitive accuracy, especially in optimizing text-image similarity tasks.

\textbf{Common space embedding.} To calculate the similarity between 3D objects and prompts, the team employs feature vector embedding in a shared space. Since the feature vectors of 3D objects and prompts have different dimensions, they utilize two MLP networks with two layers, where the output layer has the same number of units, to align them in the common vector space. They incorporate a Dropout layer~\cite{Hinton-2012} to mitigate overfitting in each network. In this shared space, the similarity between two embedding vectors, denoted as $\bf{u}$ and $\bf{v}$, is computed using the cosine similarity metric:

\begin{equation}
    \text{sim}\left (\bf{u}, \bf{v} \right ) = \frac{\bf{u} \cdot \bf{v}}{\left \| \bf{u} \right \| \left \| \bf{v} \right \|}
    \label{eq:cosine}
\end{equation}

\textbf{Loss function.} The team employs the Normalized Temperature-scaled Cross Entropy Loss (NT-Xent)~\cite{Chen-ICML2020} as the contrastive loss function. For a mini-batch of $2N$ samples $\{\bf{x}_i\}$, consisting of $N$ objects and $N$ queries, they denote $\bf{z}_i$ as the embedding vector of sample $\bf{x}_i$ in the common vector space. Let $P_i$ be the set of indices of samples that are similar to $\bf{x}_i$ in the current mini-batch, excluding $i$, \ie, $(\bf{x}_i, \bf{x}_j)$ forms a positive pair for $j \in P_i$. It should be noted that $\bf{x}_i$ can belong to multiple positive pairs, such as when two 3D objects are similar to the same query. The loss function for a positive pair $(\bf{x}_i, \bf{x}_j)$ is defined as:
\begin{equation}
    l_{i,j} = -\log\frac{\exp\left(\text{sim}\left(\mathbf{z}_{i}, \mathbf{z}_{j}\right)/\tau\right)}{\sum^{2N}_{k=1} \mathbb{I}_{[k\neq{i}, k \notin P_i]}\exp\left(\text{sim}\left(\mathbf{z}_{i}, \mathbf{z}_{k}\right)/\tau\right)}.
\end{equation}

\textbf{Training phase.} To train the proposed network, they used AdamW~\cite{Loshchilov-2017} optimizer, along with the \texttt{StepLR}, to reduce the learning during the training process. They also applied the $k$-fold cross-validation technique with $k = 5$.

\textbf{Retrieval phase.} They ensemble the results of models trained on $k$-fold by majority vote. The similarity between a 3D object and a prompt is the max value of the similarity score computed by the five models.

\begin{figure*}[!t]
    \centering
     \includegraphics[width=\linewidth]{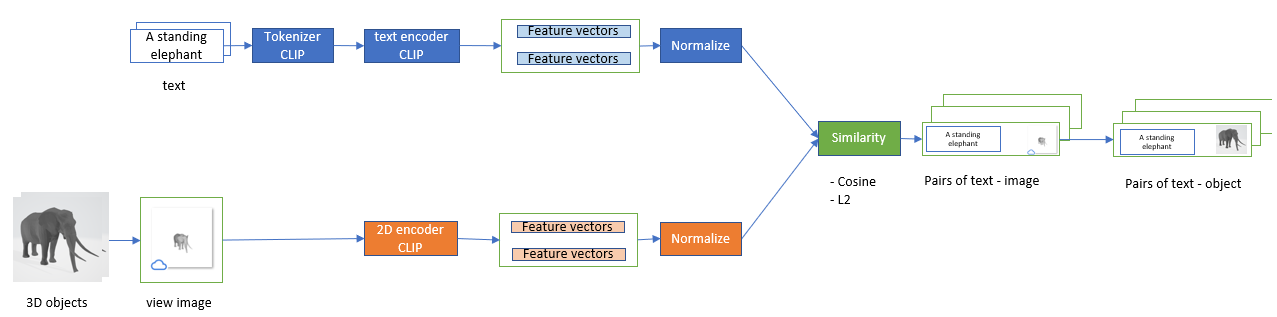}
     \caption{Proposed framework of Etinifni team.}
     \label{fig:overview_etinifni}
\end{figure*}

\highlight{\subsubsection{Data Augmentation}}

\highlight{They find that there are animal models in the ANIMAR dataset with similar appearance (\eg, animals in near families, different granularity versions of a model). Therefore, they aim to group and describe these models by corresponding descriptions to increase the training data.}

\textbf{Query clustering.} They find that the ANIMAR dataset has the following characteristics: (1) most animals in near families have a similar appearance, (2) an animal can have many 3D models with different densities of point clouds corresponding to the granularity. From these remarks, they cluster 3D models using KMeans~\cite{Lloyd-TIF1982} for further processing. Since there is nothing specific information about the objects yet, they take a naive approach that each object is represented by statistics on the set of points such as mean, variance, percentile, min, and max. 

\begin{figure}[!t]
	\centering
\includegraphics[width=\linewidth]{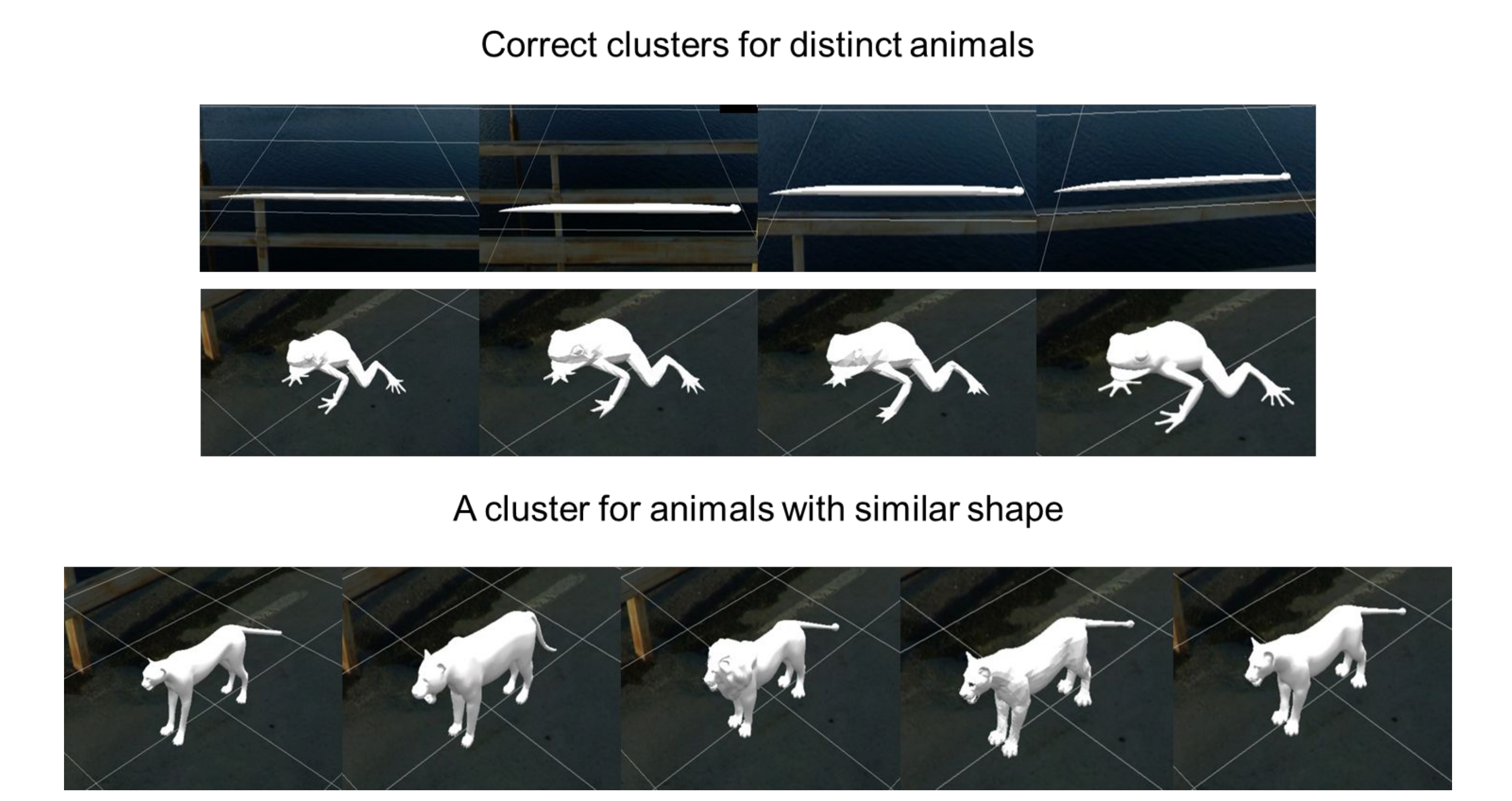}
        \caption{Clustering results are divided into two groups. The first group is the correct clusters of a specific animal (\eg, the first-row cluster is for a snake, and the second-row cluster is for fog), while the second one is clusters of animals with similar shapes (the third-row cluster for both lions and leopards).}
	\label{fig:final_clustering}
\end{figure}

The team also observed that specific local regions on the animal could aid in distinguishing it from other objects. Therefore, instead of computing statistics globally on the entire point cloud, they divided it into smaller clouds along each dimension and performed local statistics on each cloud. Specifically, they divided the height into five parts, the length into six parts, and the width into two parts. Through experimentation, they found that using 150 clusters yielded satisfactory results. The final clustering result was divided into two groups. The first group clustered distinct animals with their different versions. In contrast, the second group grouped animals with similar shapes (e.g., a group of rhinos and hippos or a group of lions and leopards), as shown in Fig.~\ref{fig:final_clustering}. \highlight{This step allowed them to manually separate these classes for further data augmentation purposes to increase the training data.}

\textbf{Descriptive query generation.} They observe that 3D animal models are mostly context-independent, except for a few insignificant cases like an arched angry cat. Based on this observation, they \textit{extract descriptive phrases} that contain the characteristics and species names of the corresponding animals. Notably, this information is present in the subject of the sentence (refer to Fig.~\ref{fig:text_analysis}). By doing so, they can convert animal descriptions into new text queries, which directs the model's attention toward these specific features.

Directly using the initial training set is insufficient to train the model due to three main reasons: (1) the number of text queries for model training is too small, about 350 prompts over 711 models, (2) adding context makes the training process more difficult, (3) the number of given text queries is insufficiently distributed for different versions of an identical animal. Therefore, through the clustering step, they utilize queries from a 3D model to assign them to its other versions. As a result, after this step, they are able to increase the original dataset by nearly three times (1100 prompts).


\subsection{Etinifni Team}
\label{sec:team_etinifni}

\subsubsection{Text-Image Learning Framework}

They develop a text-image learning framework for the text-based 3D animal fine-grained retrieval (see Fig.~\ref{fig:overview_etinifni}), including an image feature extractor and a text feature extractor.

\textbf{Image feature extraction.} For each 3D object, they extract two views from different angles and convert the text-object retrieval task into a text-image retrieval task (\ie, retrieving each 3D object by retrieving its corresponding two images). They also find that using many views extracted from 3D objects (\eg, 12 views as in the work of Su \etal~\cite{Su-ICCV2015}) is inefficient in retrieval tasks as various views from an object may cause noise and harm the model.

\begin{figure*}[!t]
     \centering
     \includegraphics[width=0.9\linewidth]{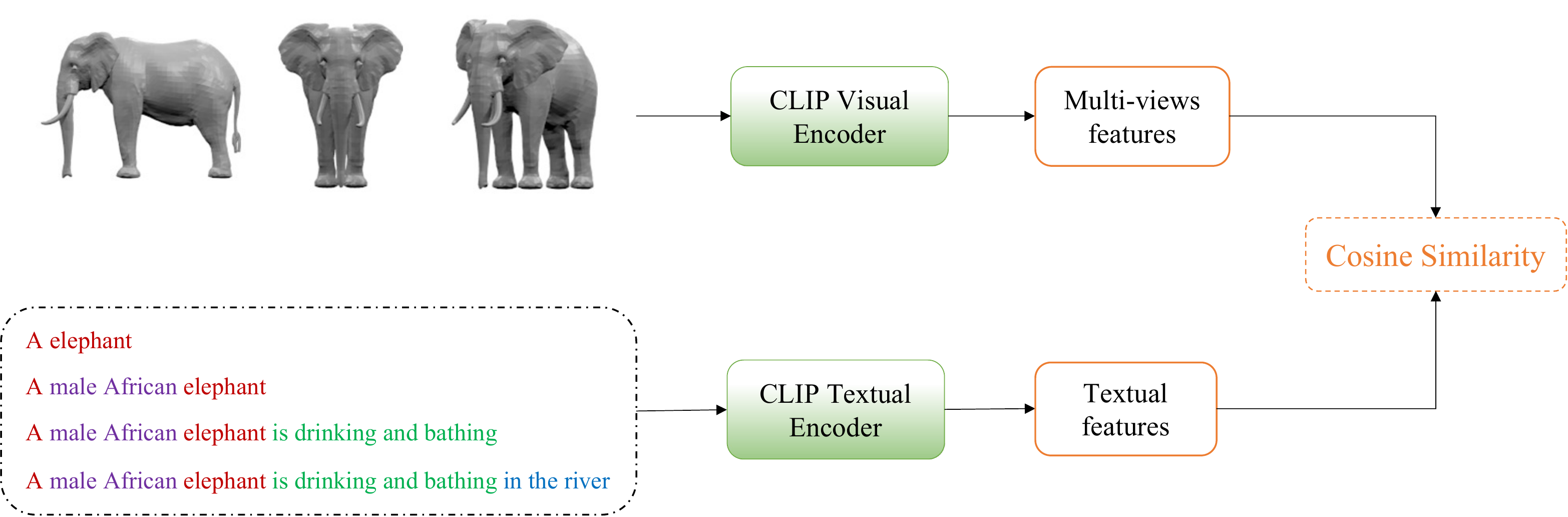}
     \caption{Proposed framework of THP team.}
     \label{fig:overview_thp}
\end{figure*}

\begin{figure}[!t]
    \centering
    \includegraphics[width=0.3\linewidth]{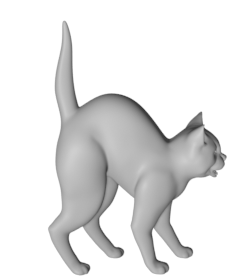}
    \includegraphics[width=0.25\linewidth]{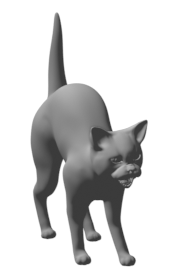}
    \caption{Two views of a cat are used in the proposed framework of the Etinifni Team.}
    \label{fig:cat}
    \vspace{-2mm}
\end{figure}

Blender \footnote{\url{https://www.blender.org}} is used to set up a camera at an appropriate distance, height, and orientation to ensure comprehensive coverage of the object's surface. Additionally, light is positioned at the camera position to provide suitable illumination. They use the horizontal view and oblique angle view of the object (depicted in Fig.~\ref{fig:cat}). CLIP image encoder~\cite{Radford-ICML2021} is used to extract features from view images. They are normalized and then used for the training stage.

\textbf{Text feature extraction.} They first pre-process text by cleaning and fixing error prompts manually. The prompts are then fed into the text tokenizer and encoder of the CLIP model~\cite{Radford-ICML2021} with backbone ViT-B16 to produce feature vectors. After that, these feature vectors are normalized and then used for the training stage.

\textbf{Loss function.} Since each text in the dataset can be paired with multiple view images, inspired by the work of Tran \etal~\cite{Tran-ICCMI2022}, they readjust the CLIP model's original loss function as suitable to the newly generated dataset, as follows:
\begin{equation}
    S_T=\frac{T \cdot T^{\tau}}{\|T\| \|T^{\tau}\|}; 
    S_I=\frac{I \cdot I^{\tau}}{\|I\| \|I^{\tau}\|};
\end{equation}
\begin{equation}
    Logits=\frac{T \cdot I^{\tau}}{\|T\| \|I^{\tau}\|};
    Target = \sigma(c \cdot \frac{S_T+S_I}{2}),
\end{equation}
where T is the text embedding matrix, and I is the image embedding matrix. From T and I, they calculate text similarity $S_T$ and image similarity $S_I$ using the cosine similarity function. The pairwise similarity between text and image, which are $Logits$, are aimed to match the mean self-similarity (of text and image) $Target$ using cross-entropy loss. $\sigma(x)$ is the softmax function and $c$ is the logit scale.

\textbf{Training phase.} They split the dataset into two subsets, the training and validation datasets, with the ratio $80\%$ and $20\%$, respectively. They trained the model for 100 epochs with a batch size of 48. They also used the early stopping technique, in which after 10 epochs, the training process is terminated if the best loss on the validation dataset does not update. Corresponding to each best loss, they obtained the weight of the model. They applied the AdamW optimization algorithm with $\beta = (0.9, 0.98)$, $\epsilon = 1\mathrm{e}{-6}$, and a learning rate of $1\mathrm{e}{-6}$. They also applied the learning rate decay technique when training.
 
\textbf{Retrieval phase.} After obtaining the similarity score of each text's correspondence to all views retrieved from corresponding 3D objects, they calculate the similarity score of each piece of text's correspondence to all 3D objects by summing the similarity score of each piece of text to two views of each 3D object as follows:
\begin{equation}
    \cos{(T_i, O_j)} = \cos{(T_i, I_{j1})} + \cos{(T_i, I_{j2})},
\end{equation}
where $cos(T_i,O_i )$ is the similarity score between text $T_i$ and 3D object $O_j$ which is calculated by the cosine similarity function. $cos(T_i,I_{j1} )$ is the similarity score between text $T_i$ and the first image $I_{j1}$ of object $O_j$ which is calculated by cosine similarity function. From the similarity score, they rank them in descending order and extract the corresponding IDs of 3D objects.


\begin{figure*}[!t]
     \centering
     \includegraphics[width=\linewidth]{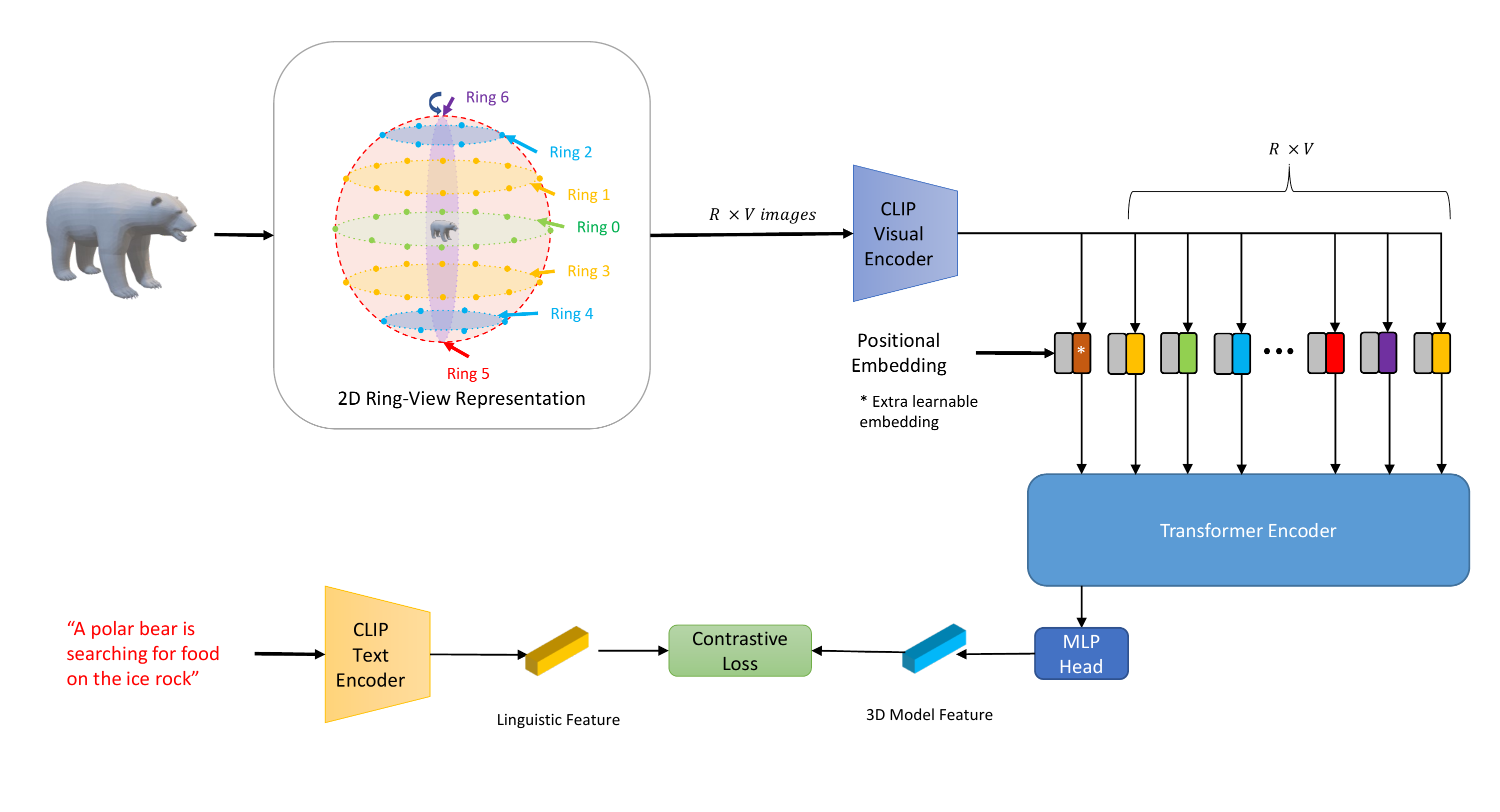}
     \caption{Proposed framework of Nero team.}
     \label{fig:overview_nero}
\end{figure*}

\subsection{THP Team}
\label{sec:team_thp}

\subsubsection{Proposed Solution}

Figure \ref{fig:overview_thp} depicts their proposed solution, in which the pre-trained CLIP model~\cite{Radford-ICML2021} is used to extract visual and textual features for each query. CLIP visual feature vectors and textual query vectors are calculated and matched using the cosine similarity function. For each view of the 3D object, they sum the six highest similarity scores. Then the object score is calculated as the maximum score of four views; each object score corresponds to four sentences. They take the average of these scores to get the final result.

\textbf{2D Projection.} They use four camera setups to take multiple views of 3D objects:

\begin{itemize}
    \item For the first camera setup, assuming the 3D object is initially solid along the $z$-axis, the camera is solid on the $Oxy$ plane and looks at the center of the object. The camera is moved around the subject to create 12 views from a distance of 30 degrees each time. 

    \item For the second camera setup, the camera is raised to 30 degrees above the $Oxy$ plane and moved around to create the next 12 views. 

    \item For the third camera setup, the camera is placed on the $Oyz$ plane and looks at the object's center. The camera moves like the first setup to create the next 12 views. 

    \item Camera for the last setup is raised from the third setup to 30 degrees to create the next 12 views. 
\end{itemize}

A total of 48 views are captured for each object. Despite generating images from other directions, their tests have shown that these 48 views provide sufficient information to observe the object's characteristics. By using Max Pooling to select the final features, additional views do not reduce the model's accuracy.

\subsubsection{Data Pre-processing}

Images are cropped and resized to 224x224 with padding of 5 pixels. For each text query, they split the sentence into small components, such as article, adjective, noun, verb, and object, and then recombine them to make different sentences (see Fig.~\ref{fig:overview_thp}). This helps the model increase the ability to recognize detailed descriptions in sentences. Nouns act as global features and detailed features are shown through adjectives and verbs. Contextual information seems to have little impact because the image has no background.


\subsection{Nero Team}
\label{sec:team_nero}

\subsubsection{Proposed Network}

Figure \ref{fig:overview_nero} shows the proposed network, containing four significant components: 2D ring-view representation, CLIP visual encoder, CLIP text encoder, and ring-view transformer encoder.

\textbf{Ring-view encoder for 3D models.} To utilize the given 3D object, they extract 2D snapshots from cameras orbiting around it. They first determine the smallest spherical hull of the object and divide it into a fixed set of $R$ latitudes, called \textit{rings}. The camera is then positioned at $V$ evenly spaced positions, facing the center of the object, which they refer to as \textit{views}. This results in $R \times V$ 2D images called ring-view images. For this work, they set $V = 12$ and $R$ ranges from $0$ to $6$, representing the cameras on the equator and the 30/60/90 latitudes from both hemispheres.

\textbf{CLIP visual encoder.} To encode the $R \times V$ 2D images generated by the ring-view method, they leverage the CLIP Visual Encoder to obtain $R \times V$ ring-view features of the 3D animal model views. The CLIP Visual Encoder is a pre-trained deep neural network that encodes natural images into high-dimensional feature vectors based on training on a large corpus of image-text pairs. Using this pre-trained visual encoder, they can effectively capture the relevant visual features of the 3D animal model without requiring additional training or fine-tuning. These encoded visual features are then aggregated for use in conjunction with the textual embeddings to perform fine-grained retrieval tasks. The use of the CLIP Visual Encoder enables us to leverage pre-existing, state-of-the-art visual representations to achieve high accuracy in 3D animal model retrieval.

\begin{table*}[!t]
    \caption{Leaderboard results of TextANIMAR competition. Best run results on the public test.}
    \label{tab:text-public}
    \centering
    \begin{tabular}{|l|c|c|c|c|c|c|c|}
    \hline
    \multicolumn{1}{|c|}{\textbf{Team}} &
      \multicolumn{1}{c|}{\textbf{NN}} &
      \multicolumn{1}{c|}{\textbf{P@10}} &
      \multicolumn{1}{c|}{\textbf{NDCG}} &
      \multicolumn{1}{c|}{\textbf{mAP}} &
      \multicolumn{1}{c|}{\textbf{FT}} &
      \multicolumn{1}{c|}{\textbf{ST}} &
      \multicolumn{1}{c|}{\textbf{FR}} \\ \hline
    \rowcolor{lightgray} TikTorch & \textbf{0.520 (1)} & 0.220 (2)  & \textbf{0.651 (1)} & \textbf{0.527 (1)} & \textbf{0.450 (1)} & \textbf{0.570 (1)} & 0.0110 (2) \\ 
    Etinifni & 0.400 (2) & \textbf{0.236 (1)}  & 0.628 (2) & 0.482 (2) & 0.370 (2) & 0.500 (2) & \textbf{0.0109 (1)} \\ 
    THP      & 0.280 (3) & 0.192 (3) & 0.541 (3) & 0.380 (3) & 0.300 (3) & 0.430 (3) & 0.0114 (3) \\ 
    Nero     & 0.080 (4) & 0.084 (4)  & 0.383 (4) & 0.168 (4) & 0.140 (4) & 0.180 (4) & 0.0130 (4) \\ 
    Polars   & 0.040 (5) & 0.032 (5)  & 0.255 (5) & 0.070 (5) & 0.000 (5) & 0.010 (5) & 0.0141 (5) \\ \hline
    \end{tabular}%
\end{table*}

\begin{table*}[!t]
    \caption{Leaderboard results of TextANIMAR competition. Best run results on the private test.}
    \label{tab:text-private}
    \centering
    \begin{tabular}{|l|c|c|c|c|c|c|c|}
    \hline
    \multicolumn{1}{|c|}{\textbf{Team}} &
      \multicolumn{1}{c|}{\textbf{NN}} &
      \multicolumn{1}{c|}{\textbf{P@10}} &
      \multicolumn{1}{c|}{\textbf{NDCG}} &
      \multicolumn{1}{c|}{\textbf{mAP}} &
      \multicolumn{1}{c|}{\textbf{FT}} &
      \multicolumn{1}{c|}{\textbf{ST}} &
      \multicolumn{1}{c|}{\textbf{FR}} \\ \hline
    \rowcolor{lightgray} TikTorch & \textbf{0.460 (1)} & \textbf{0.238 (1)}  & \textbf{0.647 (1)} & \textbf{0.525 (1)} & \textbf{0.440 (1)} & \textbf{0.585 (1)} & \textbf{0.0108 (1)} \\ 
    Etinifni & 0.360 (2) & 0.200 (2)  & 0.612 (2) & 0.460 (2) & 0.300 (3) & 0.425 (3) & 0.0114 (2) \\ 
    THP      & 0.280 (3) & 0.182 (3)  & 0.549 (3) & 0.386 (3) & 0.305 (2) & 0.430 (2) & 0.0116 (3) \\ 
    Nero     & 0.100 (4) & 0.098 (4)  & 0.398 (4) & 0.183 (4) & 0.145 (4) & 0.210 (4) & 0.0128 (4) \\ 
    Polars   & 0.040 (5) & 0.018 (5) & 0.252 (5) & 0.060 (5) & 0.015 (5) & 0.030 (5) & 0.0139 (5) \\ \hline
    \end{tabular}%
\end{table*}

\textbf{CLIP text encoder.} In their method for 3D animal model retrieval, they adopt the CLIP Text Encoder model~\cite{Radford-ICML2021} to generate textual embeddings for the descriptions of the 3D models. These embeddings are then combined with visual embeddings for fine-grained retrieval tasks. The powerful semantic representation capabilities of the CLIP Text Encoder enable us to achieve better performance in distinguishing between visually similar but semantically distinct classes of 3D animal models without requiring additional training.

\textbf{Vision transformer for ring-view features.} To capture the global information of 3D animal models, they need to gather the ring-view visual features effectively. Each ring-view feature contains the 3D model at a different angle, so they leverage the robust Transformer architecture, which is highly effective for a wide range of natural language processing and computer vision tasks. Specifically, they inherit from the Vision Transformer and consider the Ring-View extraction a patch embedding, where each ring-view is treated as a separate patch. First, the position embeddings are added to the ring-view embeddings to retain positional information. They then use the Transformer encoder to process these patches and generate a final pooled embedding for the 3D animal model. The Transformer encoder allows the model to capture long-range dependencies between the ring views and extract high-level features relevant for fine-grained retrieval. 

\textbf{Loss function.} During the training process, they employ a variant of contrastive loss called InfoNCE~\cite{Oord-2018}. They compute the InfoNCE loss function not only for the prompt and its corresponding 3D animal model but also for the prompt and other prompts, as well as for pairs of 3D animal models. For each training data batch, they randomly sample a set of prompts along with their corresponding 3D animal models. Subsequently, they compute the InfoNCE loss for the following pairs: prompt and corresponding 3D animal model, prompt and randomly sampled prompt, 3D animal model and randomly sampled 3D animal model. By considering all of these pairs, they encourage the model to learn meaningful representations that capture both the similarity and dissimilarity relationships between prompts and 3D animal models, as well as among different 3D animal models. This approach yields improved retrieval performance and more robust representations of the 3D animal models.

\subsubsection{Retrieval Phase}

To retrieve 3D animal models using text descriptions, they first encode the textual descriptions using the CLIP Text Encoder, resulting in a high-dimensional textual embedding. They then calculate the similarity scores between the textual embedding of the query and each of the final features of available 3D models using a similarity metric such as cosine similarity. The 3D models are then sorted by their similarity scores in descending order, and all the models are returned in this sorted order. This approach allows us to retrieve all relevant 3D animal models based on natural language descriptions, ranked by their similarity to the query. By leveraging the powerful semantic representation capabilities of the CLIP Model, they can achieve high accuracy in the text-based retrieval of 3D animal models.

\subsubsection{Training Details}

They used the PyTorch framework to train the model. Specifically, they used the AdamW optimizer~\cite{Loshchilov-2017} with a learning rate of $0.0001$ to optimize the model parameters. During training, they randomly sampled a set of prompts and their corresponding 3D animal models for each batch of training data. They trained the model for 100 epochs, using a batch size of 16, with the learning rate scheduled to decrease by a factor of $0.1$ at epochs $50$ and $75$. 


\section{Results and Discussions}
\label{sec:results}

The TextANIMAR track evaluates submissions on two subsets: the public and private tests. The private test comprises 50 text queries, leading to 188 query-model pairs. Besides, half of the private tests (25 text queries) are randomly selected and assigned to the public test subset to ensure fairness and prevent cheating. The leaderboard for the private test is unveiled only after the challenge's conclusion.

The leaderboard results for the public and private tests are presented in Tables~\ref{tab:text-public} and ~\ref{tab:text-private}, respectively. It is important to note that only the top-performing runs submitted by each team are displayed. However, for a fair comparison, we focus on analyzing the results from the private test, which assessed all submitted text queries.

Table~\ref{tab:text-private} illustrates the performance of the submitted methods, with the TikTorch team consistently emerging as the top-performing approach. They achieved a significant lead over other teams in all performance metrics, including the public test results (\cf~Table~\ref{tab:text-public}). Following closely, Etinifni secured the second position, with THP closely behind. Both of these teams utilized CLIP encoders for image and text representation. CLIP has shown promising performance in open-world tasks for 2D images, and its transferability to 3D point clouds is evident in their results. \highlightx{It is interesting that these two teams swap positions on FT and ST metrics, although the results are approximate.} In the public test, TikTorch and Etinifni also obtained the top two rankings, with Etinifni surpassing TikTorch \highlightx{in terms of $P@10$ and FR}. The remaining three teams, THP, Nero, and Polars, consistently performed in private and public test sets. \highlightx{Hence, these results illustrate the challenges of our ANIMAR dataset when participants did not perform excellently (the best NN result is less than 0.5, and P@10 results do not surpass 0.25). It also suggests that there is still room for improvement in addressing this research problem.}

\begin{figure}[!t]
    \centering
    \includegraphics[width=\columnwidth]{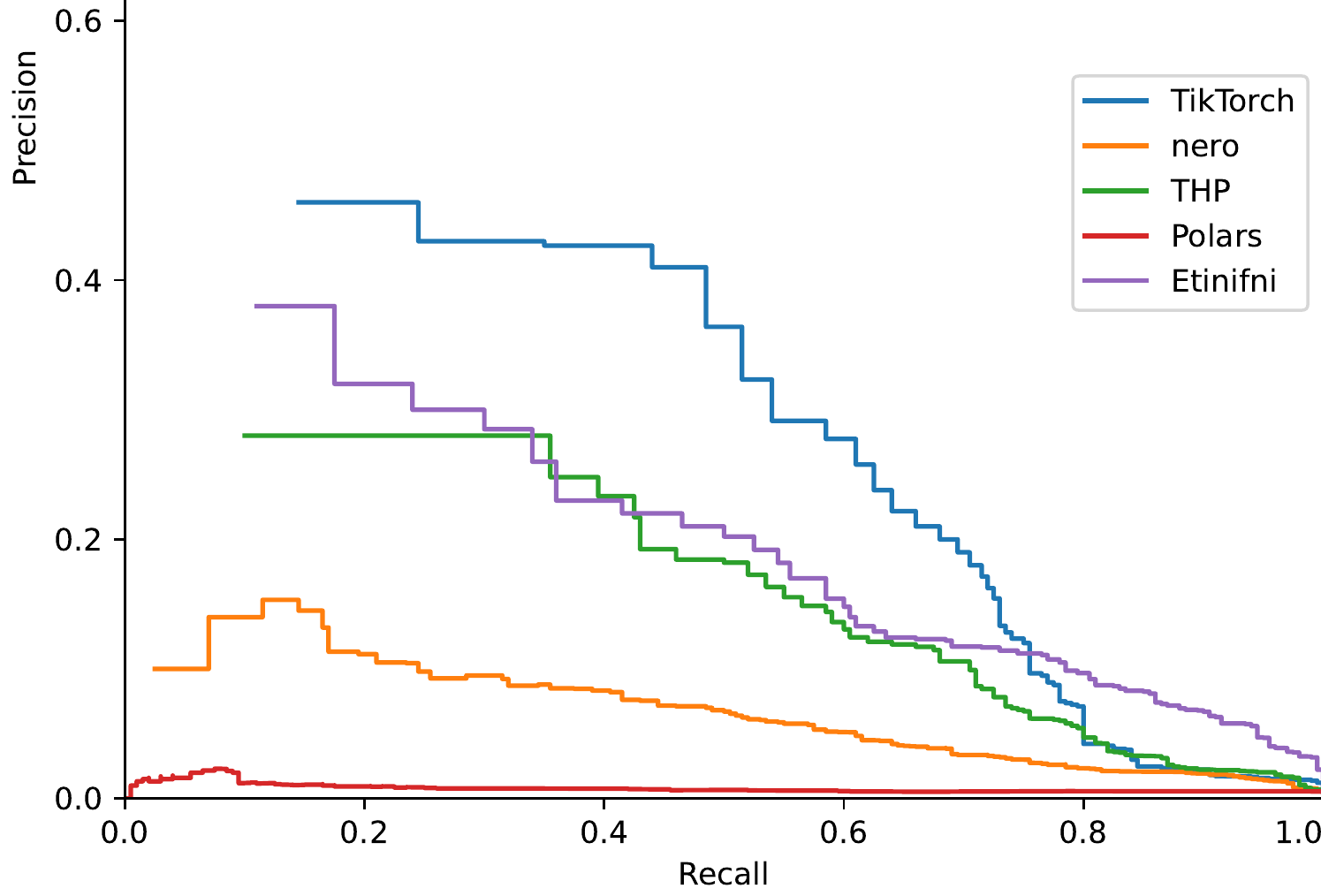}
    \caption{The visualization of precision-recall curves of submissions on the private test of teams. It can be seen that the TikTorch team achieves the best average performance with the highest area under the curve, while Etinifni obtains the highest precision at a high recall (recall $\ge 0.8$) among the five teams.}
    \label{fig:pr_curves}
    \vspace{-2mm}
\end{figure}

Figure~\ref{fig:pr_curves} displays the precision-recall curves of the submissions from five teams, namely TikTorch, Etinifni, THP, Nero, and Polars, on the private test. The graph clearly shows that TikTorch achieved the highest average performance, as indicated by the largest area under the curve. When examining the precision at a low recall threshold (recall $\le 0.4$), the three teams, TikTorch, Etinifni, and THP, demonstrate relatively acceptable precision levels. However, at a high recall threshold (recall $\ge 0.8$), Etinifni stands out as the most effective team among the five, exhibiting the highest precision.

To sum up, the view-based learning technique emerged as a successful approach for achieving high performance. It can be explained that 3D objects in the ANIMAR dataset have high-density point clouds leading to difficulty for feature extraction models~\cite{Qi-CVPR2017, Muzahid-JAS2020}. It is worth noting that these models generally randomly sample \highlightx{pointclouds} (\eg, 1024). In contrast, the utilization of view images captured by moving the trajectory camera, as shown in Fig.~\ref{fig:object2multiview}, facilitates feature learning by leveraging the semantic information and representation of 3D objects. This further confirms the efficacy of the view-based learning approach in 3D object retrieval.

\section{Conclusion}
\label{sec:conclusion}

This paper introduces a novel track for text-based retrieval of fine-grained 3D animal models along with a newly constructed ANIMAR dataset to complement existing content-based 3D object retrieval tasks. Our SHREC 2023 challenge track is designed to simulate real-life scenarios and has the potential to become a significant research direction in the field of 3D object retrieval. Despite being more challenging than previous iterations, five groups successfully participated in the track, submitting 114 runs of their proposed methods. The evaluated results of this track were satisfactory, but they also revealed the difficulties of the task at hand.

In the future, we will expand the dataset by collecting a more diverse set of 3D animal models that cover a more comprehensive range of species, postures, and environmental contexts. This expansion will enhance the generalization capability of potential solutions and improve performance on unseen 3D animal models. We also intend to generate synthetic data and \highlightx{texture maps} to augment the existing 3D animal models with different postures, backgrounds, and patterns, enabling the training of more robust and effective representation models. Another focus of our future research is investigating language models for effective text query analysis. This analysis can lead to improved retrieval performance and enable useful applications for users who are unable to draw sketches. By pursuing these avenues of research, we aim to enhance the state-of-the-art in 3D object retrieval and facilitate more intuitive and user-friendly interactions with these technologies.

\section*{CRediT authorship contribution statement}


\textbf{Trung-Nghia Le}: Conceptualization, Writing – review $\&$ editing, Project administration, Supervision. 
\textbf{Tam V. Nguyen}: Conceptualization,  Writing – review $\&$ editing. 
\textbf{Minh-Quan Le}: Software, Writing – review $\&$ editing. 
\textbf{Trong-Thuan Nguyen}: Software, Writing – review $\&$ editing.
\textbf{Viet-Tham Huynh}: Data curation. 
\textbf{Trong-Le Do}: Software, Investigation.
\textbf{Khanh-Duy Le}: Visualization. 
\textbf{Mai-Khiem Tran}: Data curation. 
\textbf{Nhat Hoang-Xuan}: Software. 
\textbf{Thang-Long Nguyen-Ho}: Software. 
\textbf{Vinh-Tiep Nguyen}: Conceptualization. 
\textbf{Tuong-Nghiem Diep}: Methodology, Writing – original draft.
\textbf{Khanh-Duy Ho}: Methodology, Writing – original draft.
\textbf{Xuan-Hieu Nguyen}: Methodology, Writing – original draft.
\textbf{Thien-Phuc Tran}: Methodology, Writing – original draft.
\textbf{Tuan-Anh Yang}: Methodology, Writing – original draft.
\textbf{Kim-Phat Tran}: Methodology, Writing – original draft.
\textbf{Nhu-Vinh Hoang}: Methodology, Writing – original draft.
\textbf{Minh-Quang Nguyen}: Methodology, Writing – original draft. 
\textbf{E-Ro Nguyen}: Methodology, Writing – original draft. 
\textbf{Minh-Khoi Nguyen-Nhat}: Methodology, Writing – original draft.
\textbf{Tuan-An To}: Methodology, Writing – original draft.
\textbf{Trung-Truc Huynh-Le}: Methodology, Writing – original draft.
\textbf{Nham-Tan Nguyen}: Methodology, Writing – original draft.
\textbf{Hoang-Chau Luong}: Methodology, Writing – original draft.
\textbf{Truong Hoai Phong}: Methodology, Writing – original draft. 
\textbf{Nhat-Quynh Le-Pham}: Methodology, Writing – original draft. 
\textbf{Huu-Phuc Pham}: Methodology, Writing – original draft. 
\textbf{Trong-Vu Hoang}: Methodology, Writing – original draft. 
\textbf{Quang-Binh Nguyen}: Methodology, Writing – original draft. 
\textbf{Hai-Dang Nguyen}: Methodology, Writing – original draft. 
\textbf{Akihiro Sugimoto}: Conceptualization.
\textbf{Minh-Triet Tran}: Conceptualization, Supervision, Funding acquisition, Writing - Review \& Editing.

\section*{Data availability}

\highlight{After the challenge concluded, the dataset has been made publicly available for academic purposes.}

\section*{Declaration of competing interest}

The authors declare that they have no known competing financial interests or personal relationships that could have appeared to influence the work reported in this paper.

\section*{Acknowledgments}


This work was funded by the Vingroup Innovation Foundation (VINIF.2019.DA19) and National Science Foundation Grant (NSF\#2025234). 

\bibliographystyle{cag-num-names}
\bibliography{short_bibtex}

\end{document}